\documentclass{article}
\usepackage[utf8]{inputenc}
\usepackage[T1]{fontenc}
\usepackage{amsmath,amssymb}
\usepackage{xcolor}
\usepackage{booktabs}
\usepackage{array}
\usepackage{caption}
\usepackage{hyperref} 

\title{\textbf{HomoEnsNER: Does Language Alignment Outperform Architectural Complexity in Gujarati Named Entity Recognition?}}

\author{
  Chandrakant K. Bhogayata\thanks{Retired Professor of Education.}\\
  \small Maharaja Krishnakumarsinhji Bhavnagar University, Bhavnagar (Gujarat, India)\\
  \small \texttt{chandrakantbhogayata@mkbhavuni.edu.in}
} 

\date{} 

\makeatletter
\def\@fnsymbol#1{\number#1}
\makeatother

\begin{document}

\maketitle
\renewcommand{\thefootnote}{\arabic{footnote}}
\setcounter{footnote}{1}

\begin{abstract}
Named Entity Recognition (NER) for Gujarati remains underexplored, hindered by the absence of capitalization cues, rich morphology, lexical ambiguity, and free word order. Prior ensemble work has emphasized architectural diversity by combining heterogeneous classifiers, multilingual encoders, or classical sequence models, rather than exploiting language-aligned monolingual pretraining. This study asks whether, for a low-resource, morphologically rich language like Gujarati, a homogeneous ensemble of a single monolingual encoder outperforms such architectural diversity. We propose HomoEnsNER, a homogeneous ensemble of five independently fine-tuned GujaratiBERT models combined via majority voting, evaluated against a single GujaratiBERT baseline and six heterogeneous alternatives, including combinations with MuRIL-base, MuRIL-large, IndicBERT, mBERT, BiLSTM, CRF, and a stacked BiLSTM-CRF-GujaratiBERT architecture. All eight models were trained under a consistent budget and evaluated using entity-level F1 on the Naamapadam Gujarati test split. HomoEnsNER achieved the highest F1 (0.8442), surpassing the baseline (0.8347) and every heterogeneous alternative (lowest: 0.7855), indicating that language alignment is a more effective, budget-conscious ensembling strategy than architectural complexity for low-resource Indian language NER.
\end{abstract}

\section{Introduction}

Named Entity Recognition (NER) is a foundational task in Natural Language Processing (NLP) that involves identifying named entities in text and classifying them into predefined categories such as person, location, and organization names. NER underpins a range of downstream applications, including information extraction, information retrieval, question answering, recommendation systems, machine translation, and knowledge base construction [1].

Most NER research to date has concentrated on high-resource languages such as English and several European and Asian languages (e.g., Chinese, Japanese, and Korean). In contrast, Gujarati NER has received far less attention. Gujarati is an Indo-Aryan language with a script derived from Devanagari. It is spoken mainly in the Indian state of Gujarat and is one of India's 22 official languages. Despite its wide use, only a handful of studies have examined NER for Gujarati.

This scarcity is closely tied to the broader low-resource status of Gujarati and most other Indian languages relative to English. Researchers working on Gujarati NER face several language-specific and resource-related challenges [2], [3], [4]:

\begin{enumerate}
  \item Unlike English, where capitalization offers a strong surface-level cue for named entities, Gujarati lacks capitalization entirely, removing a signal that many NER systems rely on.
  \item Gujarati exhibits comparatively free word order, which complicates the modeling of entity boundaries and contextual dependencies.
  \item Many Gujarati proper names double as common nouns in the lexicon, introducing lexical ambiguity that is difficult to resolve without strong contextual modeling.
  \item Gujarati is morphologically rich. Root nouns frequently fuse with postpositions, case markers, or gender/number/tense suffixes which fragments surface forms and increases sparsity for entity recognition models.
  \item Gujarati, like most Indian languages, is under-resourced: annotated corpora, POS taggers, morphological analyzers, and other NLP resources needed to support and evaluate NER systems remain limited.
\end{enumerate}

While NER has been studied extensively for some Indian languages, notably Hindi and Bengali, comparatively little work has addressed Gujarati. This gap motivates the present study.

Prior NER research has broadly progressed through rule-based systems, statistical machine learning methods (e.g., Conditional Random Fields, Hidden Markov Models, Support Vector Machines), and deep learning architectures (e.g., Bidirectional Long Short-Term Memory networks), with hybrid approaches combining elements of these directions also explored throughout [5]. More recently, two trends have come to dominate the field: (1) transformer-based Bidirectional Encoder Representations from Transformers (BERT) models [6], [7], and (2) Ensemble Learning (EL) for NER, in which entity-boundary and category predictions from multiple base learners are aggregated through a consensus mechanism to produce a single output that outperforms any individual constituent model [8].

A further consideration in adapting these trends to low-resource settings is the choice between multilingual and monolingual pretrained models. Multilingual models exploit large-scale, cross-lingual data to benefit low-resource languages, and several such options exist for Gujarati, including mBERT [7], the multilingual variant released alongside BERT itself, as well as XLM-RoBERTa [10], IndicBERT [11], and MuRIL [12]. However, prior work suggests that, for a given target language and task, monolingual models tend to outperform their multilingual counterparts because their representations are tightly aligned with the linguistic properties of that specific language [9]. Based on this evidence, we adopt the GujaratiBERT\footnote{\url{https://huggingface.co/l3cube-pune/gujarati-bert}} [13] model, pretrained on a monolingual Gujarati corpus, as the core building block of our approach because it best fits our needs, rather than relying on multilingual alternatives.

This raises a natural question for ensemble design in low-resource NER: when assembling multiple models to push performance beyond a single fine-tuned baseline, is it more effective to combine multiple instances of a single, language-aligned model (a homogeneous ensemble), or to introduce architectural diversity by combining the language-aligned model with structurally different components, such as other pretrained encoders (e.g., MuRIL, IndicBERT, mBERT), classical sequence models (e.g., CRF, BiLSTM), or additional neural layers stacked on top of it? The former uses language alignment as its primary source of reliability; the latter trades some of that alignment for architectural complexity in the hope of capturing complementary error patterns.

To investigate this question, we propose HomoEnsNER, a homogeneous ensemble learning framework for Gujarati NER built entirely from independently fine-tuned GujaratiBERT models, and we systematically compare it against a single fine-tuned GujaratiBERT baseline and a range of heterogeneous alternatives. These include ensembles that combine GujaratiBERT with other multilingual or monolingual BERT-style models (MuRIL-base, MuRIL-large, IndicBERT, mBERT), ensembles that combine GujaratiBERT with classical sequence labeling components (CRF, BiLSTM), and architectures that stack additional CRF and BiLSTM layers directly on top of GujaratiBERT. All experiments are conducted on the Gujarati split of the Naamapadam\footnote{\url{https://huggingface.co/datasets/ai4bharat/naamapadam/viewer/gu}} [14] dataset using various voting mechanisms drawn from established ensemble learning techniques, allowing us to assess the relative contribution of language alignment versus architectural complexity under consistent experimental conditions.

This study makes three main contributions: (i) while ensemble approaches to Gujarati NER have been explored using classical architectures such as BiLSTM [15], this is, to the best of our knowledge, one of the first studies to systematically investigate BERT-based ensemble learning for the task; (ii) it directly contrasts homogeneous, language-aligned ensembling against several forms of architectural heterogeneity within a single, controlled experimental setup; and (iii) it offers practical guidance for building NER systems in low-resource Indian languages, where computational budgets and the availability of multiple strong pretrained models are often limited.

\section{Related Work}

\subsection{Named entity recognition for Gujarati and Indian languages}

Named Entity Recognition (NER) for Indian languages has evolved through three broad stages: rule-based systems, statistical machine learning approaches, and transformer-based deep learning models. Despite substantial progress for languages such as Hindi, Bengali, and Telugu, Gujarati remains comparatively underexplored because of limited annotated corpora and the scarcity of language-specific NLP resources. Comprehensive reviews consistently identify Gujarati as one of the least studied Indian languages for NER while highlighting persistent challenges such as the absence of capitalization, rich morphology, lexical ambiguity, and relatively free word order. These linguistic characteristics make Gujarati NER considerably more difficult than English NER.

Early Gujarati NER systems primarily relied on statistical learning method and handcrafted linguistic rules. Garg et al. [16] developed one of the first Conditional Random Field (CRF)-based Gujarati NER systems, demonstrating that probabilistic sequence labeling substantially outperformed purely rule-based methods and achieved an F1-score of 0.832 for person, organization, and location recognition. Shah and Bhadka [17] proposed a rule-based Gujarati NER system that defined manually constructed linguistic rules to identify person, location, and organization entities, reporting approximately 70\% accuracy on their evaluation dataset. Although rule-based methods demonstrated the feasibility of Gujarati NER, they required extensive manual effort and exhibited limited generalizability to new domains.

Recently, Vora et al. [18] employed the spaCy framework to construct a Gujarati NER model together with newly created Gujarati datasets, illustrating the growing adoption of modern NLP toolkits for Gujarati language processing.

The availability of large-scale annotated datasets has significantly accelerated NER research for Indian languages. The Naamapadam corpus [14] constitutes the largest publicly available multilingual NER dataset for major Indic languages and provides standardized training and evaluation resources for Gujarati together with ten other languages. Alongside the dataset, its authors introduced IndicNER, a multilingual IndicBERT [11] model fine-tuned on Naamapadam, establishing a common standard for future research on Indian-language NER.

\subsection{Transformer-based language models for low-resource Indian languages}

Transformer-based pretrained language models have fundamentally changed the state of the art for Named Entity Recognition. Bidirectional Encoder Representations from Transformers (BERT) demonstrated that large-scale self-supervised pretraining can substantially improve downstream sequence labeling tasks through contextualized word representations. This success has motivated the development of multilingual transformer models specifically designed for Indian languages, including mBERT [7], the multilingual variant released alongside BERT itself, IndicBERT [11], and MuRIL [12], all of which have demonstrated competitive performance across multiple multilingual NLP benchmarks.

Nevertheless, recent research suggests that multilingual pretraining is not always the optimal strategy for individual low-resource languages. Joshi [9] observed that when sufficient monolingual corpora are available, language-specific BERT models consistently outperform multilingual alternatives because their representational capacity is devoted entirely to modeling a single language rather than being distributed across many. Based on these findings, monolingual transformer models, including GujaratiBERT [13], were released for several Indic languages.

Recent work on Indian-language NER shows that fine-tuning pretrained transformer models consistently outperforms traditional statistical and recurrent neural network architectures. Bahad et al. [19] demonstrated that multilingual transformer models can be effectively fine-tuned for multiple Indian languages and highlighted recurring challenges across the Indic NER landscape, such as free word order, the absence of capitalization, and morphological richness. Together, these studies position transformer-based language models as the current foundation for Indian-language NER.

\subsection{Ensemble learning for named entity recognition}

Ensemble learning seeks to improve prediction accuracy by aggregating multiple models whose errors are only partially correlated. Within NER, ensemble approaches have become increasingly popular because they generally improve reliability and reduce prediction variance compared with individual models.

Earlier research on Indian-language NER primarily combined heterogeneous statistical classifiers. Ekbal and Saha [4] proposed a multiobjective optimization framework that combined Maximum Entropy, CRF, and Support Vector Machine classifiers through optimized weighted voting, demonstrating that carefully designed classifier ensembles outperform individual learners across several Indian languages. Similar ideas were later explored for Hindi NER [20], [21], where ensemble strategies successfully combined multiple classifiers to improve entity recognition accuracy over single-model baselines.

With the emergence of deep learning, ensemble research shifted toward combining transformer encoders with other neural architectures. Recent studies have investigated combinations of transformers with CRF, BiLSTM, and other sequence-labeling models, while others aggregate predictions from multiple pretrained language models or multiple large language models. EL4NER [8], for example, proposes a multi-stage ensemble of several small-parameter language models through task decomposition and self-validation mechanisms, whereas HybridNER [22] integrates span-based neural models with large language models to improve cross-domain rigorousness.

A recent Gujarati study by Patel and Prajapati [15] found that a hard-voting ensemble of multiple BiLSTM models with FastText embeddings outperformed mBERT and a single BiLSTM baseline on the Gujarati portion of the Naamapadam [14] dataset. This remains the closest prior work to the present study, although it is related to recurrent neural architectures using word embeddings rather than fine-tuned transformer encoders.

Table 1 summarizes representative studies on Gujarati and Indian-language NER by two factual dimensions: whether an ensemble is used and, if so, whether its component models are homogeneous (multiple instances of the same encoder) or heterogeneous (different architectures or encoders); and whether the underlying language representation is monolingual or multilingual. As the table shows, every prior ensemble study in this line of work is either heterogeneous, multilingual, or both; no existing study combines a homogeneous ensembling strategy with a monolingual, language-aligned encoder.

\subsection{Research gap}

The reviewed literature shows three important gaps. First, despite recent advances, Gujarati NER remains less studied than NER for other major Indian languages. Few studies have looked at modern transformer models. Second, existing research on ensembles mainly emphasizes architectural diversity as the key benefit. It brings together different statistical learners, combines transformers with CRF or BiLSTM layers, or merges predictions from various pretrained language models rather than using homogeneous ensembles of a single language-specific model. Third, as Table 1 indicates, no previous study has systematically investigated whether maintaining language alignment through homogeneous ensembling of a monolingual transformer offers more benefits than architectural diversity. The only earlier homogeneous ensemble for Gujarati uses static FastText embeddings, not a fine-tuned monolingual transformer encoder. Thus, we explore whether a simple ensemble of language-aligned monolingual encoders can outperform more complex or diverse options in a controlled experiment.

\begin{table}[htbp]
  \centering
  \setlength{\tabcolsep}{4pt}
  \caption{Comparison of representative studies on Gujarati and Indian-language NER}
  \label{tab:related_work}
  \begin{tabular}{p{2.8cm} p{1.6cm} p{2.2cm} p{2.3cm} p{2.5cm} p{2.8cm}}
    \hline
    \textbf{Study} & \textbf{Language} & \textbf{Primary Model} & \textbf{Ensemble Type} & \textbf{Encoder Alignment} & \textbf{Research Focus} \\
    \hline
    Garg et al. [16] & Gujarati & CRF & None & Monolingual & Statistical NER \\
    Shah \& Bhadka [17] & Gujarati & Rule-based & None & Monolingual & Rule-based NER \\
    Vora et al. [18] & Gujarati & spaCy & None & Monolingual & Toolkit-based NER \\
    Naamapadam / IndicNER [14] & Indic & IndicNER & None & Multilingual & Benchmark dataset + model \\
    Bahad et al. [19] & Indian & Fine-tuned transformers & None & Multilingual & Cross-lingual fine-tuning \\
    Ekbal \& Saha [4] & Bengali, Hindi, Telugu & CRF+ME+SVM & Heterogeneous & Monolingual & Classifier ensemble \\
    Patel \& Prajapati [15] & Gujarati & BiLSTM ensemble & Homogeneous & Monolingual (FastText) & Embedding-based ensemble \\
    EL4NER [8] & General & Multiple LLMs & Heterogeneous & Multilingual & Multi-LLM ensemble \\
    HybridNER [22] & General & Hybrid neural+LLM & Heterogeneous & Multilingual & Span+LLM hybrid \\
    HomoEnsNER (this study) & Gujarati & GujaratiBERT & Homogeneous & Monolingual (Transformer) & Language-aligned ensemble \\
    \hline
  \end{tabular}
\end{table}

\section{Methodology}

This section describes the dataset, the base encoder, the baseline fine-tuning procedure, and the seven ensemble and hybrid configurations evaluated in this study. All configurations are built on top of GujaratiBERT and are grouped into three families that reflect the central question of this paper: (i) a homogeneous ensemble of independently fine-tuned GujaratiBERT instances, (ii) heterogeneous ensembles that combine GujaratiBERT with other pretrained encoders or classical sequence models, and (iii) stacked architectures that add classical layers directly on top of GujaratiBERT. Section 3.5 describes the aggregation (voting) mechanisms used across the ensembles, and Section 3.6 reports the hyperparameter settings for all eight configurations.

\subsection{Dataset}

We conducted all experiments on the Gujarati partition of the Naamapadam dataset [14], the largest publicly available multilingual NER corpus for Indic languages including Gujarati. The training split has been automatically created using the Samanantar parallel corpus by projecting automatically tagged entities from an English sentence to corresponding Indian language translation. The authors of [14] also have created human annotated testsets of Gujarati and other eight Indian languages. The train, validation, and test splits of the dataset contain sentence-level examples annotated with entity tags or labels following the BIO tagging scheme. The tag or label set includes PER, ORG, and LOC for Person, Organization, and Location names, respectively. We use the standard Naamapadam Gujarati splits as released.

Table 2 displays the programmatically computed number of sentences, tokens, and named entities in the train, validation, and test splits of the Gujarati partition of the Naamapadam dataset.

\begin{table}[htbp]
  \centering
  \caption{Number of sentences, tokens, and named entities in the three Gujarati splits of the Naamapadam dataset}
  \label{tab:dataset_splits}
  \begin{tabular}{lccc}
    \hline
    \textbf{Split} & \textbf{Sentences} & \textbf{Tokens} & \textbf{Named Entities} \\
    \hline
    Train & 472,845 & 6,218,281 & 765,471 \\
    Validation & 2,389 & 31,501 & 3,875 \\
    Test & 1,076 & 13,617 & 1,737 \\
    \hline
  \end{tabular}
\end{table}

Table 2 shows that the train split is comparatively very large. It has 472,845 sentences, 6,218,281 tokens, and 765,471 named entities.

In Table 3, we present the tag or label distribution of the total 6,218,281 tokens of the train split.

It can be observed from Table 3 that 79.18\% of the tokens have the label O (outside of a named entity), the label B-PER (beginning of a Person name) is more frequent than the remaining named entity labels, whereas the label I-LOC (inside a Location name) is the least frequent.

\begin{table}[htbp]
  \centering
  \caption{The label distribution of the tokens of the Gujarati split of the Naamapadam dataset}
  \label{tab:label_distribution}
  \begin{tabular}{lccc}
    \hline
    \textbf{Label} & \textbf{ID} & \textbf{Token Count} & \textbf{Percentage} \\
    \hline
    O & 0 & 4,923,455 & 79.18\% \\
    B-PER & 1 & 321,674 & 5.17\% \\
    I-PER & 2 & 283,043 & 4.55\% \\
    B-ORG & 3 & 205,652 & 3.31\% \\
    I-ORG & 4 & 190,116 & 3.06\% \\
    B-LOC & 5 & 238,145 & 3.83\% \\
    I-LOC & 6 & 56,196 & 0.90\% \\
    \hline
  \end{tabular}
\end{table}

\subsection{Base encoder: GujaratiBERT}

GujaratiBERT [13] is a monolingual BERT-style encoder that was pretrained on a large Gujarati text corpus. We choose it as the base encoder for all configurations in this study because monolingual pretraining provides representations that match the target language better than multilingual alternatives [9]. This leads to the central comparison of this paper: whether this language alignment, replicated across multiple independently fine-tuned instances, is a more effective ensembling strategy than adding architectural diversity. GujaratiBERT is used as the only encoder in the homogeneous ensemble and as the single baseline, as well as one component in each of the heterogeneous ensembles discussed in Section 3.4.

\subsection{Baseline fine-tuning}

As a reference point, we fine-tuned a single GujaratiBERT model on the Gujarati Naamapadam train split for the token classification (NER) task using a linear classification head over the encoder's final hidden states. This baseline configuration corresponds to Model 1 in Table 4 and establishes the performance floor against which all ensemble and hybrid configurations are compared.

All eight configurations were trained under a single, consistent budget: a maximum of 10 epochs with early stopping (patience of 3 epochs, monitored on validation F1). This applies uniformly to the pretrained encoders (GujaratiBERT, MuRIL-base, MuRIL-large, IndicBERT, mBERT) as well as to the randomly initialized classical components (BiLSTM, CRF) in Models 4, 7, and 8.

The BiLSTM component in Models 4 and 8 operates on full GujaratiBERT embeddings as input features. The CRF component in Model 7 operates on PCA-reduced GujaratiBERT embeddings, reduced to 32 dimensions, as static input features, rather than being trained jointly with the encoder. In Model 8, by contrast, the BiLSTM and CRF layers are stacked directly on top of GujaratiBERT's output representations and trained jointly with the encoder in a single end-to-end model (Section 3.4.4).

\subsection{Ensemble and hybrid configurations}

Eight configurations were evaluated in total, including a single baseline model and seven others, organized into three families.

\subsubsection{Homogeneous Ensemble}

The core configuration of this study, HomoEnsNER, consists of five independently fine-tuned GujaratiBERT models (Model 2, Table 4). Each of the five instances was fine-tuned on the same training split using an identical architecture but a different random seed, sampled with replacement, and a learning rate sampled from the range 1.5e-05 to 3.5e-05, in order to encourage diversity in the errors made by individual models while preserving full language alignment across the ensemble. Predictions from the five models were combined using majority voting at the token or entity-span level (Section 3.5).

\subsubsection{Heterogeneous Ensembles with Other Pretrained Encoders}

Three configurations combine GujaratiBERT with other multilingual pretrained encoders: (i) GujaratiBERT + MuRIL-base + MuRIL-large (Model 3), (ii) GujaratiBERT + IndicBERT + mBERT (Model 5), and (iii) GujaratiBERT + IndicBERT (Model 6). Each component encoder was independently fine-tuned on the same Gujarati Naamapadam training split before being combined via weighted voting, except for Model 3, which uses majority voting (Table 4). These configurations test whether introducing encoders pretrained on different (typically multilingual) corpora contributes complementary error patterns that offset their weaker language alignment relative to GujaratiBERT.

\subsubsection{Heterogeneous Ensembles with Classical Sequence Models}

Two configurations combine GujaratiBERT with classical sequence labeling components trained as independent, parallel predictors rather than stacked layers: GujaratiBERT + BiLSTM + CRF (Model 4) and GujaratiBERT + CRF (Model 7). In Model 4, the BiLSTM component was trained on full GujaratiBERT embeddings as input features, and its predictions were combined with GujaratiBERT's via majority voting. In Model 7, the CRF component was trained on PCA-reduced GujaratiBERT embeddings (32 dimensions) as static input features, and its predictions were combined with GujaratiBERT's via weighted voting (Table 4).

\subsubsection{Stacked Architecture}

The final configuration (Model 8) departs from the ensemble paradigm and instead stacks a BiLSTM layer followed by a CRF layer directly on top of GujaratiBERT's output representations, forming a single end-to-end model rather than a voting-based combination of independently trained predictors. As with all other configurations, this model was trained for a maximum of 10 epochs with early stopping (patience of 3 epochs, monitored on validation F1), using random seeds sampled with replacement. This configuration serves as a point of comparison between ensembling (combining independent predictions) and architectural stacking (adding sequential-modeling capacity within a single model).

\subsection{Ensemble aggregation strategies}

Two voting mechanisms were used to combine predictions across ensemble members, following standard ensemble learning practice [8]:

\textbf{Majority voting:} each ensemble member casts an equal-weight vote for the predicted label at each token position (or entity span), and the label receiving the plurality of votes is selected. This was used for the homogeneous GujaratiBERT ensemble (Model 2), the GujaratiBERT + MuRIL-base + MuRIL-large ensemble (Model 3), and the GujaratiBERT + BiLSTM + CRF ensemble (Model 4), reflecting settings where component models are expected to have comparable reliability.

\textbf{Weighted voting:} each ensemble member's vote is scaled by a weight before aggregation. This was used for the remaining heterogeneous ensembles (Models 5--7), reflecting settings where component models differ substantially in architecture or language alignment and are not assumed to be equally reliable. Weights were determined by each component model's validation F1 score.

\subsection{Hyperparameter settings}

Table 4 summarizes the batch size, learning rate, and ensemble aggregation method used for each of the eight configurations. All eight configurations were trained for a maximum of 10 epochs with early stopping (patience of 3 epochs, monitored on validation F1).

\begin{table}[htbp]
  \centering
  \caption{Hyperparameter settings for the Gujarati NER configurations}
  \label{tab:hyperparameters}
  \small
  \begin{tabular}{cp{4.2cm}ccp{2.6cm}}
    \hline
    \textbf{No.} & \textbf{Configuration} & \textbf{Batch Size} & \textbf{Learning Rate} & \textbf{Ensemble Method} \\
    \hline
    1 & Single baseline GujaratiBERT & 16 & 2.0e-05 & Not applicable \\
    2 & Ensemble of 5 GujaratiBERT (HomoEnsNER) & 16 & 1.5e-05 to 3.5e-05 & Majority voting \\
    3 & GujaratiBERT + MuRIL-base + MuRIL-large & 16 & 2.0e-05 & Majority voting \\
    4 & GujaratiBERT + BiLSTM + CRF & 64 (BiLSTM) & 2.0e-05 / 1.0e-03 & Majority voting \\
    5 & GujaratiBERT + IndicBERT + mBERT & 16 & 2.0e-05 & Weighted voting \\
    6 & GujaratiBERT + IndicBERT & 16 & 2.0e-05 & Weighted voting \\
    7 & GujaratiBERT + CRF & 16 & 2.0e-05 & Weighted voting \\
    8 & BiLSTM + CRF stacked on GujaratiBERT & 16 & 1.0e-05 / 1.0e-03 & Not applicable \\
    \hline
  \end{tabular}
\end{table}

\subsection{Evaluation metric}

All configurations are evaluated on the held-out test split of the Gujarati Naamapadam dataset using entity-level (span-level) precision, recall, and F1-score under the BIO scheme, computed using the Python package \texttt{seqeval}. Entity-level F1 is used as the primary metric throughout this paper, consistent with standard NER evaluation practice and with the metric reported by prior work on this dataset [14], [15], to allow direct comparison.

\subsection{Implementation details}

All experiments were conducted using PyTorch with CUDA on an NVIDIA GeForce RTX 4060 GPU. Fine-tuning time varied substantially by component. All pretrained-encoder fine-tuning runs (GujaratiBERT, MuRIL-base, MuRIL-large, IndicBERT, mBERT) and the CRF-only component (Model 7) took approximately 35 to 40 minutes each. The BiLSTM component (Models 4 and 8), trained under the same 10-epoch, patience-3 budget, took 6 hours, 37 minutes, and 29 seconds, with early stopping triggered at epoch 2.

\section{Results}

Table 5 reports the entity-level F1 scores obtained on the held-out Gujarati Naamapadam test split for all eight configurations described in Section 3. For consistency with Table 4, configurations are numbered and ordered exactly as in Section 3 (Model 1 to Model 8).

\begin{table}[htbp]
  \centering
  \caption{Experimental results for Gujarati NER}
  \label{tab:results}
  \small
  \begin{tabular}{cp{4.5cm}ccc}
    \hline
    \textbf{No.} & \textbf{Configuration} & \textbf{Ensemble Method} & \textbf{F1 Score} & \textbf{$\Delta$ vs. Baseline} \\
    \hline
    1 & Single baseline GujaratiBERT model & Not applicable & 0.8347 & 0.0000 (baseline) \\
    2 & Ensemble of 5 GujaratiBERT (HomoEnsNER) & Majority voting & \textbf{0.8442} & \textbf{+0.0095 (+1.14\%)} \\
    3 & GujaratiBERT + MuRIL-base + MuRIL-large & Majority voting & 0.8322 & -0.0025 (-0.30\%) \\
    4 & GujaratiBERT + BiLSTM + CRF & Majority voting & 0.8115 & -0.0232 (-2.78\%) \\
    5 & GujaratiBERT + IndicBERT + mBERT & Weighted voting & 0.7972 & -0.0375 (-4.49\%) \\
    6 & GujaratiBERT + IndicBERT & Weighted voting & 0.7967 & -0.0380 (-4.55\%) \\
    7 & GujaratiBERT + CRF & Weighted voting & 0.8156 & -0.0191 (-2.29\%) \\
    8 & BiLSTM + CRF stacked on GujaratiBERT & Not applicable & 0.7855 & -0.0492 (-5.89\%) \\
    \hline
  \end{tabular}
\end{table}

\subsection{Overall ranking}

The homogeneous ensemble of five independently fine-tuned GujaratiBERT models (Model 2) achieved the highest F1 score of the eight configurations, at 0.8442. This is the \emph{only} configuration in the study that improved on the single fine-tuned baseline (Model 1, F1 = 0.8347); every heterogeneous ensemble and the stacked architecture scored below the baseline. Ranked from best to worst: Model 2 (0.8442) > Model 1 (0.8347) > Model 3 (0.8322) > Model 7 (0.8156) > Model 4 (0.8115) > Model 5 (0.7972) > Model 6 (0.7967) > Model 8 (0.7855).

\subsection{Homogeneous vs. heterogeneous ensembles}

The homogeneous GujaratiBERT ensemble (Model 2) outperformed the closest heterogeneous configuration, GujaratiBERT + MuRIL-base + MuRIL-large (Model 3), by 1.2 F1 points, and outperformed the weakest ensemble, GujaratiBERT + IndicBERT (Model 6), by 4.75 F1 points. Notably, Model 3, the heterogeneous configuration built entirely from other BERT-style encoders and combined via majority voting, came closest to matching the baseline, falling only 0.25 points short of it. All other heterogeneous ensembles, regardless of whether their non-GujaratiBERT components were other pretrained encoders (Models 5, 6) or classical sequence models (Models 4, 7), fell further below the baseline than Model 3 did.

\subsection{Ensembling vs. architectural stacking}

The stacked architecture (Model 8), in which BiLSTM and CRF layers were trained jointly with GujaratiBERT rather than combined through voting, produced the lowest F1 score of all eight configurations (0.7855), 4.92 points below the baseline and 5.87 points below the homogeneous ensemble. This is a larger drop than any of the ensemble-based configurations, including the two configurations that also used classical sequence components as \emph{independent} predictors (Models 4 and 7).

\subsection{Effect of voting mechanism}

The three majority-voting configurations (Models 2, 3, 4) occupy the top three ranks among the ensembles that are not Model 1 itself, while two of the three weighted-voting configurations (Models 5, 6) occupy the bottom two ranks among ensembles. The weighted-voting configuration built from other pretrained encoders alone (Model 7, GujaratiBERT + CRF) is a partial exception, outperforming the majority-voting Model 4 (GujaratiBERT + BiLSTM + CRF) despite also relying on weighted voting. This suggests that the degree of language alignment and reliability of the component models, rather than the voting mechanism per se, is the dominant factor in ensemble performance in this study.

\section{Discussion}

\subsection{Language alignment outperforms architectural complexity}

The central finding of this study is unambiguous within the scope of our experiments: preserving language alignment through homogeneous ensembling of a monolingual, language-specific encoder outperformed every attempt to introduce architectural or representational diversity. This directly answers the research question posed in the Introduction: for Gujarati NER under the experimental conditions tested here, homogeneous ensembling of a language-aligned model is a more effective strategy than heterogeneous ensembling or architectural stacking, at least at the scale and training budget explored in this study.

This result is notable because it runs counter to the general intuition motivating ensemble learning, namely that combining models with different inductive biases or error distributions should yield stronger, more complementary predictions than combining near-identical models [8]. In our experiments, the opposite pattern held: every configuration that introduced a component with a different language alignment (MuRIL, IndicBERT, mBERT) or a different architectural family (BiLSTM, CRF) performed worse than the single GujaratiBERT baseline, while five independently seeded and independently fine-tuned instances of the \emph{same} language-aligned encoder, differing only in random seed and learning rate, produced enough complementary error patterns to improve on the baseline through majority voting alone.

\subsection{Why heterogeneity hurt rather than helped}

We interpret this pattern as reflecting a trade-off between diversity and reliability. Effective ensembling depends on component models making sufficiently uncorrelated errors while individually being reasonably accurate [4], [8]. In our heterogeneous configurations, the non-GujaratiBERT components carried a systematic disadvantage: MuRIL, IndicBERT, and mBERT are multilingual encoders whose representational capacity is distributed across many languages rather than concentrated on Gujarati [9], while the classical components (BiLSTM, CRF) were trained on comparatively limited or static input representations, namely full or PCA-reduced GujaratiBERT embeddings rather than jointly optimized features. Rather than contributing complementary but comparably accurate predictions, these components appear to have introduced a weaker, less language-aligned voice into the aggregation, which weighted or majority voting was not able to fully offset.

This interpretation is consistent with the relative ranking within the heterogeneous group. Model 3 (GujaratiBERT + MuRIL-base + MuRIL-large), which combines three BERT-style transformer encoders of comparable architectural depth, came closest to the baseline. Models 5 and 6, which introduced IndicBERT and mBERT, encoders trained across a substantially larger and more diverse set of languages, degraded performance more sharply, suggesting that the degree of multilingual dilution in the component encoder, not merely its transformer architecture, matters for ensemble quality on a low-resource, morphologically rich language such as Gujarati.

\subsection{Ensembling vs. stacking}

The comparatively poor performance of the stacked architecture (Model 8) offers a complementary insight. Unlike the ensemble configurations, which combine the \emph{independent} predictions of separately trained models, Model 8 trains BiLSTM and CRF layers jointly with GujaratiBERT end-to-end. Under the fixed training budget used throughout this study (10 epochs, patience of 3), this configuration performed worst overall. One plausible explanation is optimization difficulty: the BiLSTM component required substantially longer to train even as an independent predictor (6 hours, 37 minutes, with early stopping triggered at epoch 2, versus 35--40 minutes for the pretrained encoders), suggesting that the added recurrent and CRF layers converge more slowly than the pretrained encoder they sit on top of. When trained jointly rather than independently, this mismatch in convergence behavior may have interfered with GujaratiBERT's own fine-tuning, rather than supplementing it. This suggests that, at least for Gujarati NER with a language-aligned transformer as the base encoder, stacking additional sequence-modeling capacity is less effective than either fine-tuning the encoder alone or ensembling multiple independently fine-tuned instances of it.

\subsection{Relation to prior work}

Our findings are broadly consistent with the direction reported by Patel and Prajapati [15], whose homogeneous ensemble of BiLSTM models with FastText embeddings outperformed both mBERT and a single BiLSTM baseline on the same Gujarati Naamapadam data. The present study extends that finding from static word embeddings to fine-tuned transformer encoders and shows that the advantage of homogeneous, language-aligned ensembling persists, and indeed becomes sharper, once the base representation itself is a strong, monolingual pretrained encoder rather than a static embedding. At the same time, our results complicate the picture painted by heterogeneous-ensemble successes reported outside the low-resource Indic setting, such as EL4NER [8] and HybridNER [22]. Those studies combine components that are individually strong and reliable, such as multiple capable LLMs or span models paired with LLMs. By contrast, the heterogeneous components available for Gujarati NER, such as multilingual encoders and classical sequence models, are less well aligned to the target language than GujaratiBERT itself.

This suggests that the benefit of architectural heterogeneity in ensembling may be conditional on the availability of multiple individually strong, comparably reliable components, a condition that does not hold for most low-resource Indian languages, where a single well-pretrained, monolingual encoder may be the only genuinely strong option available.

\subsection{Practical implications}

For practitioners building NER systems for Gujarati and similarly low-resourced Indian languages, these results offer a concrete, budget-conscious recommendation: rather than investing computational effort in acquiring, fine-tuning, and combining multiple different pretrained encoders or in stacking additional neural layers, it may be more effective to fine-tune multiple independent instances of a single strong, language-aligned encoder and combine their predictions through simple majority voting. This approach requires only one type of pretrained model, avoids the engineering overhead of aligning heterogeneous architectures and tokenizers, and in our experiments outperformed every heterogeneous or stacked alternative while training in a comparable or shorter amount of time than the classical sequence-model components.

\subsection{Limitations}

Several limitations qualify these conclusions. First, all experiments were conducted on a single dataset (the Gujarati split of Naamapadam) and a single language, so the generality of the homogeneous-over-heterogeneous pattern to other low-resource Indic languages or other NER corpora remains to be tested. Second, each configuration was trained once under a fixed budget (10 epochs, patience of 3); we do not report variance across multiple independent runs of each configuration, so the precision of the F1 differences in Table 5, particularly the relatively small gap between Models 1, 2, and 3, should be interpreted with some caution. Third, the voting mechanism (majority vs. weighted) was assigned by configuration family rather than tuned exhaustively for each ensemble, so it remains possible that a different aggregation strategy could narrow the gap between the homogeneous and heterogeneous configurations. Finally, the classical components (BiLSTM, CRF) were trained on frozen or PCA-reduced GujaratiBERT features rather than being allowed to co-adapt with the encoder in the ensemble settings (as distinct from the stacked setting of Model 8); a more tightly integrated training regime for these components might narrow the performance gap observed here.

\section{Conclusion}

This study set out to answer a specific question for low-resource Indian-language NER: when building an ensemble to improve on a single fine-tuned transformer, is it more effective to preserve language alignment by combining multiple instances of one monolingual encoder, or to introduce architectural diversity by combining structurally different models? Using GujaratiBERT as the base encoder and the Gujarati split of the Naamapadam dataset as a controlled testing situation, we compared a homogeneous ensemble of five independently fine-tuned GujaratiBERT models against a single baseline, three heterogeneous ensembles built with other pretrained encoders, two heterogeneous ensembles built with classical sequence models, and one architecture stacking BiLSTM and CRF layers directly on top of GujaratiBERT.

The homogeneous ensemble, HomoEnsNER, achieved the best entity-level F1 score (0.8442) among all eight configurations and was the only configuration to improve on the single fine-tuned baseline (0.8347). Every heterogeneous ensemble and the stacked architecture scored below the baseline, with the degradation generally proportional to how far the added component's language alignment or training regime departed from GujaratiBERT itself. Multilingual encoders trained across many languages hurt performance more than other monolingual-adjacent BERT-style models, and jointly trained stacked layers hurt performance more than independently trained ensemble components.

These findings suggest that, for Gujarati and plausibly for other morphologically rich, low-resource Indian languages, the reliability gained from language-aligned pretraining outweighs the complementary error patterns that architectural diversity is generally expected to provide. In practical terms, this offers a simple and computationally economical recipe for practitioners: fine-tune several independent instances of the strongest available monolingual encoder and combine them with majority voting, rather than investing effort in acquiring and aligning multiple heterogeneous architectures or encoders.

At the same time, this conclusion is drawn from a single dataset, a single language, and single training runs per configuration, so it should be treated as an initial, controlled finding rather than a general law of ensembling in low-resource NLP. Future work could test whether this pattern holds across other Indic languages and datasets, examine the effect of increasing ensemble size beyond five members, explore tuned or learned voting mechanisms rather than fixed majority or weighted schemes, and assess variance across multiple independent runs of each configuration. Extending this comparison to other low-resource language families, and to tasks beyond NER, would help clarify how broadly the advantage of language-aligned homogeneous ensembling generalizes beyond the Gujarati setting studied here.

\section*{Declaration}

During the preparation of this manuscript, we used Grammarly and Claude AI tools for grammar correction, language polishing, and improving sentence clarity. After using the tools, we reviewed and edited the content as necessary and take full responsibility for the content of the publication.

\end{document}